\documentclass[letterpaper, 10 pt, conference]{ieeeconf}  
\usepackage{graphicx}
\usepackage{times}
\usepackage{epsfig}
\usepackage{amsmath}
\usepackage{amssymb}
\newcommand\netname{TOTEM}
\usepackage{subcaption}
\usepackage{amsfonts}
\usepackage{booktabs}
\usepackage{siunitx}
\usepackage{arydshln}
\usepackage{url}

\IEEEoverridecommandlockouts                              

\title{\LARGE \bf
Transparent Object Tracking with Enhanced Fusion Module }

\author{
    Kalyan Garigapati\textsuperscript{1}\quad Erik Blasch\textsuperscript{2}\quad Jie Wei\textsuperscript{3}\quad Haibin Ling\textsuperscript{1} \vspace{0.2cm}\\ 
    \textsuperscript{1}Stony Brook University, Stony Brook, NY, USA \\
    \textsuperscript{2}Air Force Office of Scientific Research, Arlington, VA, USA \\
    \textsuperscript{3}City College of New York, New York, NY, USA\\
    \texttt{\{kgarigapati, hling\}@cs.stonybrook.edu}\\
    \texttt{jwei@ccny.cuny.edu}\\
    \texttt{erik.blasch@gmail.com}
}

\begin{document}

\maketitle

\thispagestyle{plain}
\pagestyle{plain}

\begin{abstract}
    Accurate tracking of transparent objects, such as glasses, plays a critical role in many robotic tasks such as robot-assisted living. Due to the adaptive and often reflective texture of such objects, traditional tracking algorithms that rely on general-purpose learned features suffer from reduced performance.  Recent research has proposed to instill transparency awareness into existing general object trackers by fusing purpose-built features. However, with the existing fusion techniques, the addition of new features causes a change in the latent space making it impossible to incorporate transparency awareness on trackers with fixed latent spaces. For example, many of the current days' transformer-based trackers are fully pre-trained and are sensitive to any latent space perturbations. In this paper, we present a new feature fusion technique that integrates transparency information into a fixed feature space, enabling its use in a broader range of trackers. Our proposed fusion module, composed of a transformer encoder and an MLP module, leverages key query-based transformations to embed the transparency information into the tracking pipeline. We also present a new two-step training strategy for our fusion module to effectively merge transparency features. We propose a new tracker architecture that uses our fusion techniques to achieve superior results for transparent object tracking. Our proposed method achieves competitive results with state-of-the-art trackers on TOTB, which is the largest transparent object tracking benchmark recently released. Our results and the implementation of code will be made publicly available at \url{https://github.com/kalyan0510/TOTEM}.

\end{abstract}

\section{INTRODUCTION}

Object tracking is a fundamental problem in robotics that aims to locate and identify an object in a sequence of images or videos. Researchers have dedicated much effort to addressing various challenges in generic object tracking \cite{tomp, rts, dimp, Fan&Ling21iros, transt, keeptrack}, which is mostly concerned with opaque objects. However, tracking of transparent objects is a somewhat less explored topic. Transparent objects, such as glass and plastic, are common in everyday life, and reliably tracking them has numerous practical applications in robotics \cite{ClearGrasp, Weng_2020}, surveillance, and augmented reality. Transparent object tracking can be used in robotic medical procedures to track and visualize the movement of glass vials and syringes.


Though there is a pressing need to track transparent objects reliably, it is very challenging. These objects possess unique properties since they primarily borrow texture from the background and are also reflective. When such object moves, its appearance changes drastically due to background influence. These properties pose a severe issue to appearance-based deep trackers as they tend to extract feature information from visual cues of striking color and edge patterns. Thereby generic trackers tend to rely on falsely extracted background features, thus performing poorly on transparent objects.
\begin{figure}[t!]
\includegraphics[width=\linewidth]{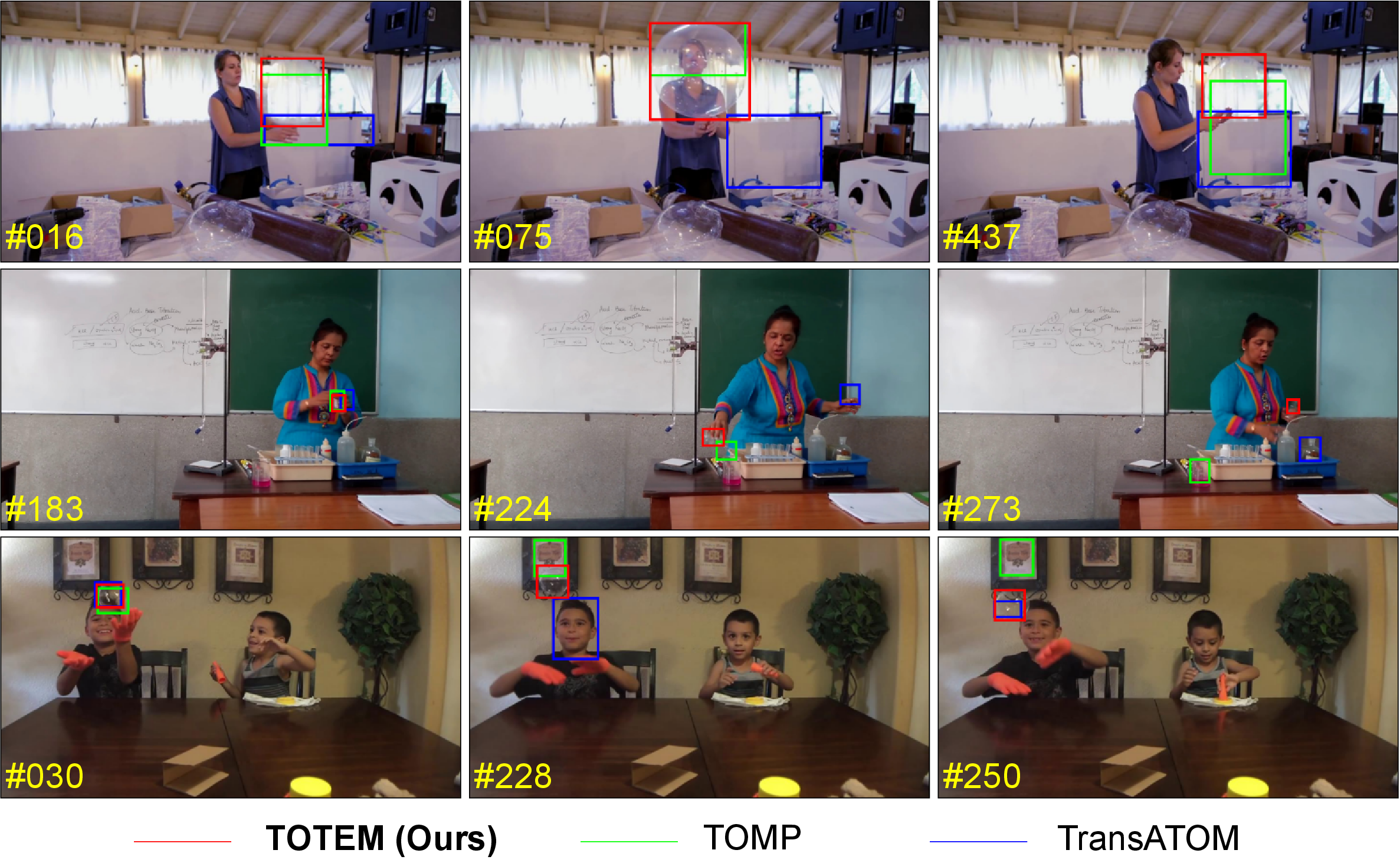}
\caption[ Qualitative results  ]
        {\small  Qualitative comparison of the proposed TOTEM tracking algorithm with state-of-the-arts \cite{tomp, totb} on three challenging sequences from TOTB \cite{totb}. Owing to the effective fusion technique tailored for transparency awareness, TOTEM can accurately localize transparent objects under challenging scenarios. \textit{All figures in this paper are best viewed digitally, in color, significantly zoomed in.}}
        
\label{fig:topright}

\end{figure}

In contrast to some application-specific tracking tasks such as people tracking \cite{long2018tracking} or UAV tracking \cite{li2016multi}, transparent object tracking suffers from the absence of a dedicated training dataset. Consequently, end-to-end training to improve tracking performance is impractical currently.  To overcome this challenge, recent research has proposed to use knowledge transfer techniques to imbue generic trackers with transparency awareness. Specifically, features from a backbone module trained for transparent object segmentation are fused into the tracker pipeline. It is hypothesized that such a backbone encodes transparent textures well and thus helps trackers perform with accuracy.

However, while the above feature fusion approach seems promising, it is not always straightforward. Simple fusion techniques may not always be effective, as the fusion of features in a pipeline can disrupt the feature space and require retraining of the entire model to learn to utilize the fused features. Retraining can be particularly challenging when labeled data is scarce. In \cite{totb}, Fan et al. chose to use ATOM \cite{atom} and DiMP \cite{dimp} trackers, which are capable of consuming fused features without requiring full retraining, as they consist of fully online-learned modules. However, this approach may not be viable for many state-of-the-art trackers that rely on components pre-trained on large datasets.

Our proposed fusion technique selectively fuses transparency features with the original ones without disrupting the feature space, thus allowing for integration with most trackers. Our module consists of a transformer encoder block and an MLP block. The transformer block has attention layers to efficiently fuse transparency information. The MLP block projects the fused features back into the original feature space. This property of our fusion module allows for integration of learned transparency priors in many trackers.  

Moreover, we have demonstrated that the fusion module can be trained efficiently in a two-step process. Specifically, an additional pre-training step is performed, which compels the fusion module to rely exclusively on transparency features for tracking by cutting off the feed of originally extracted features to the fusion module. Further, we design a new tracker called TOTEM (Transparent Object Tracking with feature Enhancing Module) that uses our fusion methods to achieve robust performance on transparent object scenarios, as shown in Fig. \ref{fig:topright}. 

\vspace{2mm}
The contributions of this work are as follows: 
\begin{itemize}
\setlength
\item We propose a novel transparency feature fusion module for tracking transparent objects. 
\item We devise a novel two-step training strategy for effective learning.
\item We design a new tracker architecture TOTEM aimed at better transparent object tracking.
\item We perform extensive experiments over the transparent object tracking benchmark TOTB~\cite{totb} and perform ablation studies to showcase the benefit of our design choices.
\end{itemize}


\section{RELATED WORK}

\vspace{1mm}\noindent\textbf{Transparent objects and tracking.}   
Transparent objects present unique challenges for classification, segmentation, and tracking due to their optical properties. Previous studies \cite{delpozo2007detecting, fleming2005low, xu2015transcut} have proposed handcrafted techniques that rely on reflective and refractive light properties to model transparent objects. In recent years, due to the progress of machine learning techniques, algorithms that gain complex skills by learning from huge data have showed promising results. The works of \cite{trans2seg, translab} prove that learnable components such as convolution-based feature extractors and transformer encoder blocks can leverage from training on labelled transparent object datasets for accurate segmentation. Similarly \cite{zhu2021transfusion, DBLP:journals/corr/abs-1803-04636} utilizes learning over huge data to model transparent objects.


However, the problem of tracking transparent objects remains a challenge due to the scarcity of labeled datasets. To address this, Fan et al. \cite{totb} constructed a large tracking benchmark dedicated to transparent objects. Further, they proposed a transfer learning approach that introduces transparency awareness into existing generic object trackers. However, their method is only applicable to trackers with online learned tracking modules. In contrast, our proposed fusion module does not have any restrictions on applicability. Given the recent popularity of transformers in tracking architectures \cite{tomp, aiatrack, stark}, which are typically pre-trained models, our approach shows promise in leveraging these strong baselines. Particularly, our model is built on top of TOMP \cite{tomp}, a transformer model prediction tracker.




\vspace{1mm}\noindent\textbf{Segmentation and Dataset.}
Research over transparent objects has gained momentum in recent years, with several datasets such as \cite{DBLP:journals/corr/abs-1803-04636, translab, trans2seg} providing valuable sources for learning transparency priors for object segmentation. In this work, we leverage the pixel-level segmentation dataset \cite{trans2seg}, which includes annotations for five different categories of transparent objects. This dataset closely represents real-world transparent objects and provides accurate pixel-level labeling for improved localization. While the dataset from \cite{trans2k} offers exhaustive labeling, it is not used in this work due to the synthetic nature of the objects and their limited representation of real-world scenarios. Further, we use different portions of the \textit{Transparent Object Tracking Benchmark} (TOTB) dataset \cite{totb} for training and benchmarking our tracker algorithm.


\vspace{1mm}\noindent \textbf{Feature fusion.} 
Recently, more attention has been devoted to multi-modal architectures. These works mainly benefit from the fusion techniques \cite{ding2015robust, kart2018depth, DBLP:journals/corr/abs-2007-02041} like concatenation \cite{totb}, feature pruning \cite{gao2019deep}, and re-weighting \cite{zhu2018fanet, liu2020robust}. These fusion methods mainly aim at merging the information from multiple modalities and do not necessarily operate as learnable modules. Lately, more robust fusion methods were proposed that utilize the transformer's attention mechanism to fuse features. For example, the works of \cite{shvetsova2022everything, tomp} use transformers for fusing image features. 

\begin{figure*}
\definecolor{x}{RGB}{255, 203, 163}
\definecolor{xdash}{RGB}{204,229,255}
\definecolor{xdashdash}{RGB}{255,229,153}

\includegraphics[width=\linewidth]{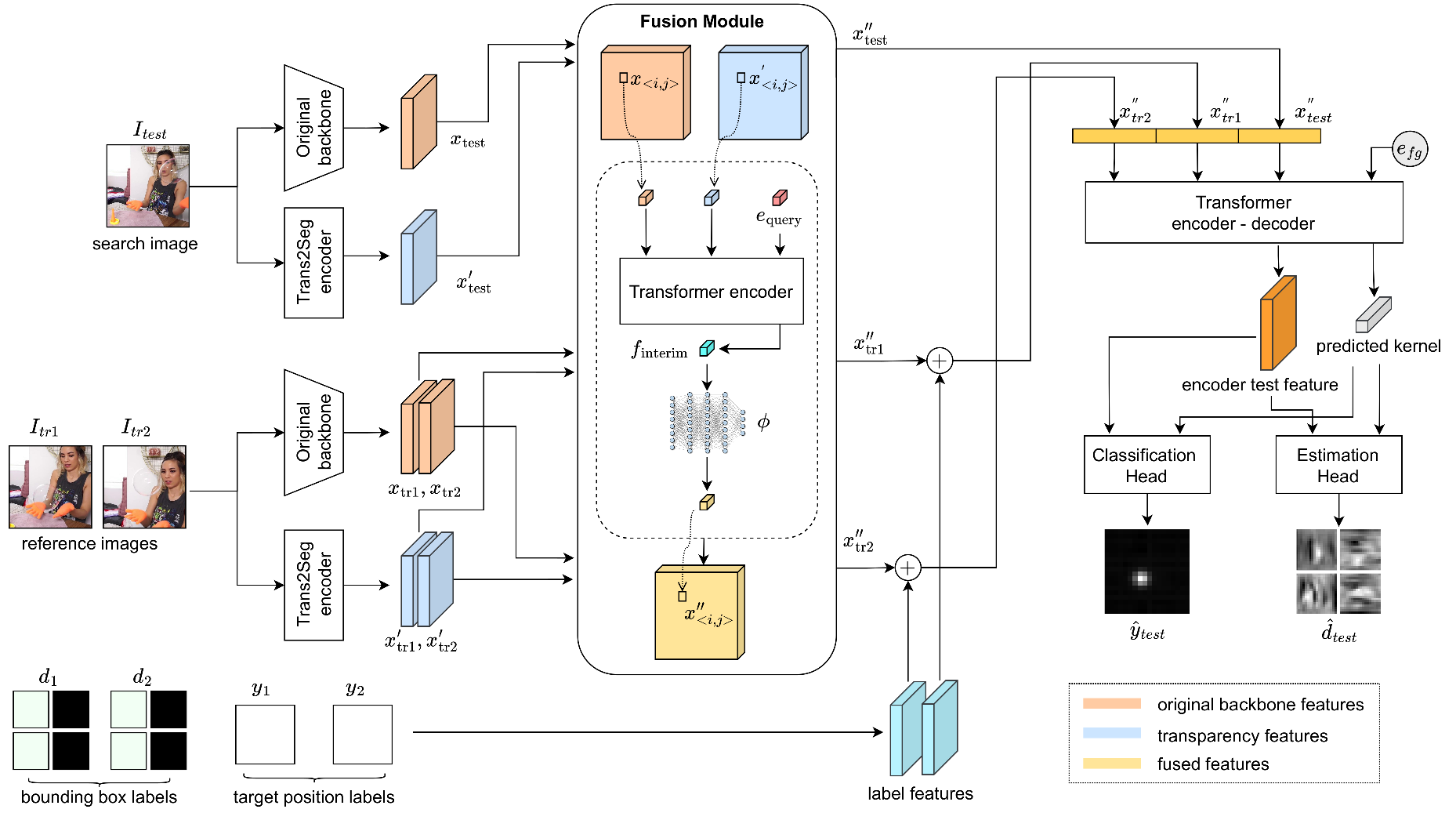}
\caption[ Full architecture overview of our tracking pipeline  ]
        {\small  Full architecture overview of our tracking pipeline. The original ResNet backbone extracts the features \(x_i\) [\fcolorbox{x}{x}{\rule{0pt}{4pt}\rule{4pt}{0pt}}] for each input image in test and train frames. The Trans2Seg backbone extracts the transparency features \(x'_{i}\) [\fcolorbox{xdash}{xdash}{\rule{0pt}{4pt}\rule{4pt}{0pt}}]. These features are fused by the fusion module to produce enriched features \(x''_{i}\) [\fcolorbox{xdashdash}{xdashdash}{\rule{0pt}{4pt}\rule{4pt}{0pt}}]. After the fusion module, the features flow through the model predictor pipeline for gaining discriminative ability. Lastly, the predicted model is applied over the encoder features to localize the target.
        }
\label{fig:TOTEMarchitecture}
\end{figure*}

Our proposed fusion technique distinguishes itself from existing fusion methods by being designed to work with pre-trained networks. Unlike existing fusion modules, which are trained as part of the end-to-end training of the network, our fusion module is trained separately to produce features that are compatible with pre-trained networks. To achieve this, we equip our fusion module with MLPs to project the features to the known latent space of the pre-trained network. These differences make our proposed technique well-suited for our specific use case.

\section{PROPOSED METHOD}

\label{chap:architecture}

The core idea of our proposed method is to enhance the effectiveness of generic object trackers for transparent object tracking. The TOMP framework (detailed in section \ref{sec3:tomp}) serves as the baseline object tracker. Next, we describe a separate network for extracting transparency features in section \ref{sec3:trans2seg}. Then, we present a novel fusion technique in section \ref{sec3:fusion} that combines these features with the baseline object tracker to enhance its effectiveness.

\subsection{Baseline Tracker - TOMP}
\label{sec3:tomp}

One of the robust paradigms for visual object tracking is discriminative model prediction-based target localization. In this approach, a kernel (target model) is predicted/estimated to accurately represent the appearance of the target object and is used to localize the target in subsequent frames. In \cite{tomp}, Mayer et al. proposed a transformer-based model predictor TOMP that utilizes the self-attention operations between test and reference branch features to produce a kernel.


TOMP consists of a test and a training branches. The training branch operates on two input ground-truth/memory frames \(I_{\rm tr1}, I_{\rm tr2} \in \mathbb{R}^{H \times W \times 3}\) where \(H\) and \(W\) indicate the image size. In the train branch, the target state information (bounding box size and position) is encoded and fused with deep image features \(x_{\rm tr1}, x_{\rm tr2} \in \mathbb{R}^{h \times w \times c}\). Features from both the training and test branches are jointly processed in the transformer model predictor. This module contains a transformer encoder and a decoder block. The encoder produces enhanced features by reasoning across the test and train branch features. The decoder operates on the processed features from the encoder to generate the desired kernel. This kernel is then applied over the encoder features using separate classification and regression heads to produce the target's center response map \(\hat{y} \in \mathbb{R}^{h \times w \times 1}\) and the bounding box size estimations response map \(\hat{d}  \in \mathbb{R}^{h \times w \times 4}\) respectively. Here \(\hat{d}\)  represents the offsets from the predicted center point to the sides of the bounding box, encoded as (left, top, right, bottom) adjustments.

The entire network is trained end-to-end by minimizing classification loss \(\mathcal{L}_{1}\), and regression loss \(\mathcal{L}_{2}\)  produced over the input of randomly selected frame triplets \(\langle I_{\rm tr1}, I_{\rm tr2}, I_{\rm te} \rangle\) from a video sequence. 

We propose to utilize this architecture for the problem of transparent object tracking and further make modifications (see Fig. \ref{fig:TOTEMarchitecture}) to improve its performance. 
\subsection{Transparency Feature extraction}
\label{sec3:trans2seg}

As discussed earlier, one of the challenges the trackers face with transparent object tracking is the transparency property causing visual distortion over the target's appearance cues. To overcome this issue, trackers need to gain an understanding of the texture of transparent objects. Specifically, they should be able to abstract out the texture of a transparent object from its background and use this knowledge to localize the same object (even when it adapted a different background). Such ability is not something a typical opaque object tracker would gain while training. So in our work, we adopt a backbone that is trained to extract the transparency features. One simple method to benefit our tracker with transparency awareness is to train it over transparent object video sequences. Since we lack annotated video sequences, we incorporate transparency awareness by transferring it from another model (trained for a different objective).

We propose to use a separate backbone network (as shown in Fig. \ref{fig:TOTEMarchitecture}) that has the ability to understand the transparent object's texture (caused by refraction, reflection, and translucence). Motivated by the transfer learning approach in \cite{totb}, we adopt a similar approach of using the feature extractor from a segmentation network. Particularly, the segmentation network Trans2Seg~\cite{trans2seg} trained to segment and classify pixels belonging to transparent objects is used. By hypothesizing that such a segmentation network must intermediately learn to encode the patterns from transparency features like reflectivity, refractivity, and translucence, we propose to use the feature extractor and encoder part of Trans2Seg \cite{trans2seg} to produce transparency features.




The Trans2Seg segmentation network consists of a convolution-based backbone, a transformer encoder module, a transformer decoder module and a segmentation head, connected sequentially in this order. In the end-to-end training environment, the incentive of the backbone and transformer encoder would be to produce features that encode the unique properties of transparent objects. The decoder and the segmentation head would learn specific priors to categorize the transparent objects. Since we are mainly interested in image encodings, we adopt the backbone and encoder module part of Trans2Seg as the transparency feature extractor for our tracker. 

The backbone module takes in an input image \(I \in \mathbb{R}^{H \times W \times 3} \) and produces a feature vector \(\tilde{x} \in \mathbb{R}^{h \times w \times c}\), where \(H\) and \(W\) are the image height and image width respectively, and \(h\), \(w\) and \(c\) are the height, width, and the number of channels of the produced feature map. Further, the transformer encoder operates over the input feature \(\tilde{x}\) and produces a globally attended and enriched feature map \(x'\) which has the same shape as that of \(\tilde{x}\). We refer our reader to \cite{trans2seg} for more details of this module.

\subsection{Fusing the transparency features}
\label{sec3:fusion}
\noindent\textbf{Why fusion.} One way to utilize transparency features for tracking is to replace the tracker's backbone with the above transparent feature extractor directly. But this may hurt the tracking performance because the transparency backbone is trained for a less related objective and thus may not extract features specific to the tracking problem. For example, a motion-blur-affected object is never encountered when training the Trans2seg network, whereas correctly extracting motion features is critical for tracking. So, we adopt a fusion-based approach to take advantage of the transparency feature while still retaining essential cues for tracking. Also, this way, the tracker learns to selectively ingress the useful encodings of the input image detail.  


However, there are certain challenges to using transparency features in the above-discussed transformer model predictor architecture. Firstly, all the components in this tracker are offline learned, meaning that any change in architecture that modifies intermediate feature space must be accompanied by offline re-training. The perk of direct inference without training after feature fusion, as observed in \cite{totb}, does not exist with the selected baseline TOMP. Further, we do not have a large-scale training dataset consisting of transparent object video sequences. So we must adopt a simple fusion mechanism that does not require full-scale re-training from scratch.

We found that it is best to fuse the transparency features into the TOMP pipeline just before the transformer model predictor block.  This way, we can leverage the strong local and global reasoning provided by the transformer encoder-decoder module over the transparency features. 

The fusion module (depicted in Fig. \ref{fig:TOTEMarchitecture}) is designed taking into account the following constraints:
\begin{itemize}
\setlength\itemsep{0pt}
\item[-] The end-to-end model, after the transparency feature fusion, should not require re-training over the large datasets, given their lack of availability
\item[-] fusion of transparency features should not regress the tracker's performance on transparent object tracking
\item[-] it should be lightweight both in terms of the number of learnable parameters and the number of computations
\end{itemize}

To be able to reuse most of the learning modules, we designed our fusion module to be trained without having to re-train the existing components of the TOMP. While this design choice helps with the above constraints, it poses certain challenges. The TOMP model predictor is completely made of learned parameters, and it expects the input features to belong to a specific \textit{feature space}. The \textit{feature space} refers to the mapping between each channel in the feature vector and the set of specific patterns that activate a channel's response. Most of the machine-learned components are sensitive to the feature space of the input. For example, we cannot simply replace the backbone network of a classification model with a better feature extractor and see a performance improvement. At least the classification heads have to be re-trained before the model can produce any meaningful output.   

For the same reason, we cannot simply concatenate the transparency features with the features extracted by the TOMP backbone to achieve performance improvement. In fact, this will cause the network to lose performance because the transparency features are unexpected perturbations (noise) to the model predictor. So, we propose a feature fusion module and a training strategy that produce enriched features by combining useful cues from each source. Because of the training objective, the module produces a fused feature that would align with the feature space of the original TOMP backbone.

\vspace{1mm}\noindent\textbf{Fusion Module.}
Our transformer-based feature fusion module sits between the backbone and transformer encoder stages of the TOMP pipeline and fuses the features \(x_{}  \in \mathbb{R}^{h \times w \times c}\) and \(x'^{}_{} \in \mathbb{R}^{h \times w \times c}\) into a new feature \(x''^{}_{} \in \mathbb{R}^{h \times w \times c}\).

This module is designed to operate pixel-wise rather than to use global context information. So, the fusion occurs between the corresponding feature vectors \(x_{\langle i,j \rangle} \in \mathbb{R}^{c} \) and \(x'^{}_{\langle i,j \rangle} \in \mathbb{R}^{c} \) at every pixel position \(\langle i,j \rangle \in \{[0, h) \times [0, w)\}\). Note that attention operations do not occur across spatial locations. 

The module consists of two main components: 1) Transformer Encoder and 2) a Fully Connected Projection module. 

\vspace{1mm}
\subsubsection{Transformer Encoder}  
 The Transformer Encoder fuses the vectors \(x_{\langle i,j \rangle}\) and \(x'^{}_{\langle i,j \rangle}\) by transforming a query embedding \(e_\mathrm{query}\) into an intermediate feature representation \(f_\mathrm{interim}\) (shown in eq. \ref{eq:concat} and \ref{eq:TEnc}). Inspired by the architecture described in \cite{tomp, detr}, we designed this module \( T_{\rm enc}\) with multiple encoder layers. But different from \cite{detr}, we do not use a \(1 \times 1\) convolutional layer to project the features into a smaller dimension, as this would throw away important detail. Also, we do not add any positional embeddings, as no spatial information needs to be preserved. Each encoder layer follows standard architecture and consists of a multi-head self-attention module and a feed-forward network. We perform experiments in the next section exploring the effect of using a query embedding versus using one of the transformed input features. 

\begin{equation}
 z = {\rm concat}( x_{\langle i,j \rangle}, x'^{}_{\langle i,j \rangle},  e_\mathrm{query}) \in \mathbb{R}^{3 \times c}
\label{eq:concat}
\end{equation}
\begin{equation}
f_\mathrm{interim} = T_{\rm enc}(z)
\label{eq:TEnc}
\end{equation}

\subsubsection{Fully Connected Projection Module} Additionally, hypothesizing that we need a separate module \(\phi\) to project the fused features \(f_\mathrm{interim}\) onto the latent space, on which the TOMP's model predictor operates, we employ a two-layer fully connected neural network (see eq. \ref{eq:mlpmod}). Further, to match the distribution of feature activations across spatial and channel dimensions between the newly projected feature and the original feature \(x^{}_{\langle i,j \rangle}\), we add an instance normalization layer at the end of the module \(\phi\).
\begin{equation}
    x''^{}_{\langle i,j \rangle} = \phi(f_\mathrm{interim})
    \label{eq:mlpmod}
\end{equation}

\subsection{Tracker Pipeline}
\label{sec3:pipeline}
The end-to-end structure of the proposed model TOTEM is illustrated in Fig. \ref{fig:TOTEMarchitecture}. First, we use the original backbone network to extract the deep image features of both the test and train branch input frames. Parallelly, we also extract the transparency features for the same set of input frames. These features are then fused independently using the proposed fusion module. The fused features in the training branch are further combined with target state encodings. The features from both branches are flattened and concatenated to form a sequence of feature vectors. This sequence is then processed by the transformer encoder to produce enhanced features by reasoning globally across frames. Next, the Transformer Decoder predicts the target model weights by using the output of the transformer encoder. Finally, the predicted model is applied to the test branch features output by the transformer encoder to localize the target.  
 
\subsection{Training}
\label{sec3:training}

\begin{figure}
\includegraphics[width=\linewidth]{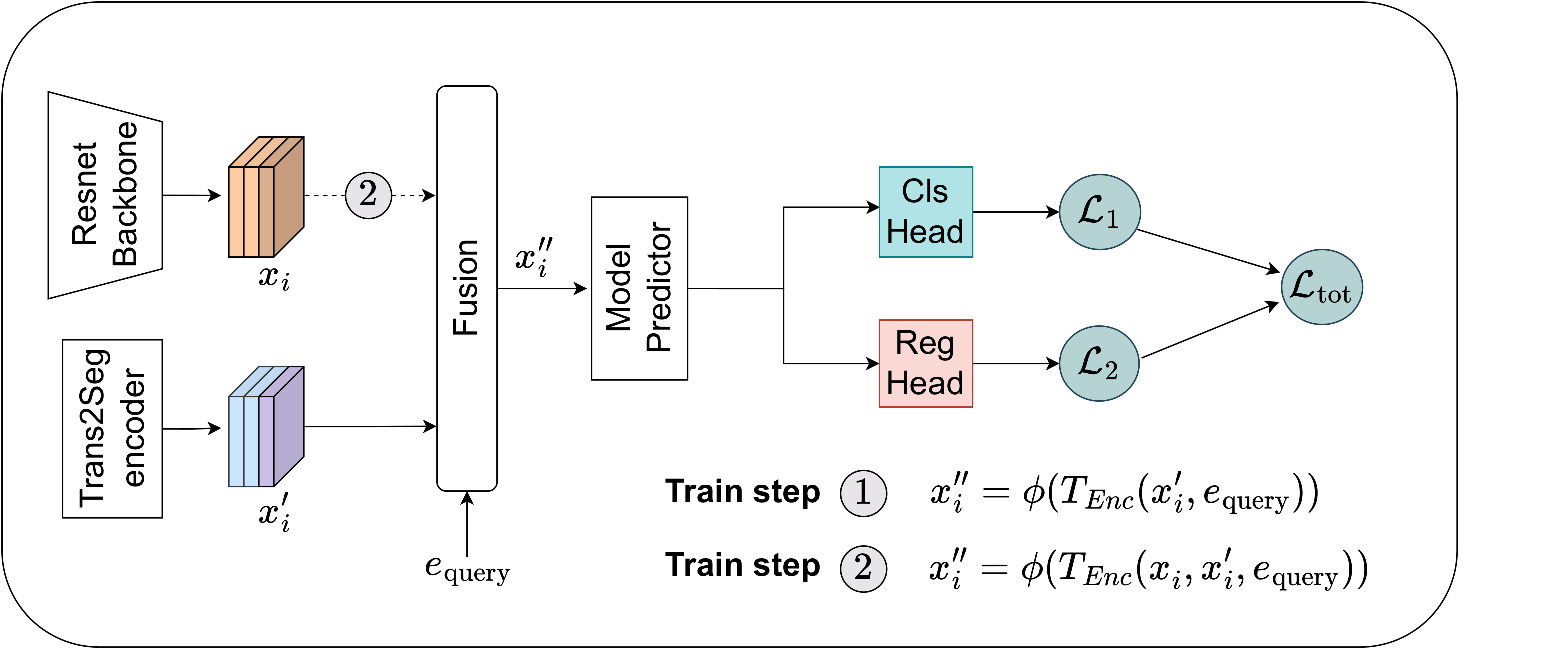}
\caption[ Two step training process ]
        {\small 
        The two step process of training the fusion module
        } 
\label{fig:twosteptrain}
\end{figure}

All the components belonging to the TOMP, and the Trans2Seg model, are initialized with pre-trained weights while the weights of the fusion module are initialized randomly following a Xavier initialization \cite{glorot2010understanding}. The training is performed over two steps where only the fusion module weights are updated with each back-propagation. In the first step (illustrated in Fig. \ref{fig:twosteptrain}), TOTEM does not use the features extracted by the original backbone. Instead, the fusion module only uses transparency features to produce a compatible output. In the first step configuration, the fusion module learns to use the transparency features (belonging to an unrelated feature space) with TOMP's model predictor. The second step follows the usual setting where both the original and transparency features are input to the fusion module. 

Empirically this two-step approach showed better performance compared to the usual training approach. We hypothesize that this approach works effectively because, without the first step, the fusion module learns to over-rely on the original features which are already in the feature space the module is learning to project into. So, by forcing the module to learn to solely use the transparency features, we encourage the fusion module to first learn to recognize transparency features. Then during the second step, it leverages the learned priors from step one to effectively fuse transparency features into the pipeline.  


\section{Experiments}
\label{chap:experiments}

\subsection{Implementation and Setup}
\label{sec4:impl}

\begin{figure*}[!ht]
        \centering
        \begin{subfigure}{0.327\textwidth}
            \centering
            \includegraphics[width=\linewidth]{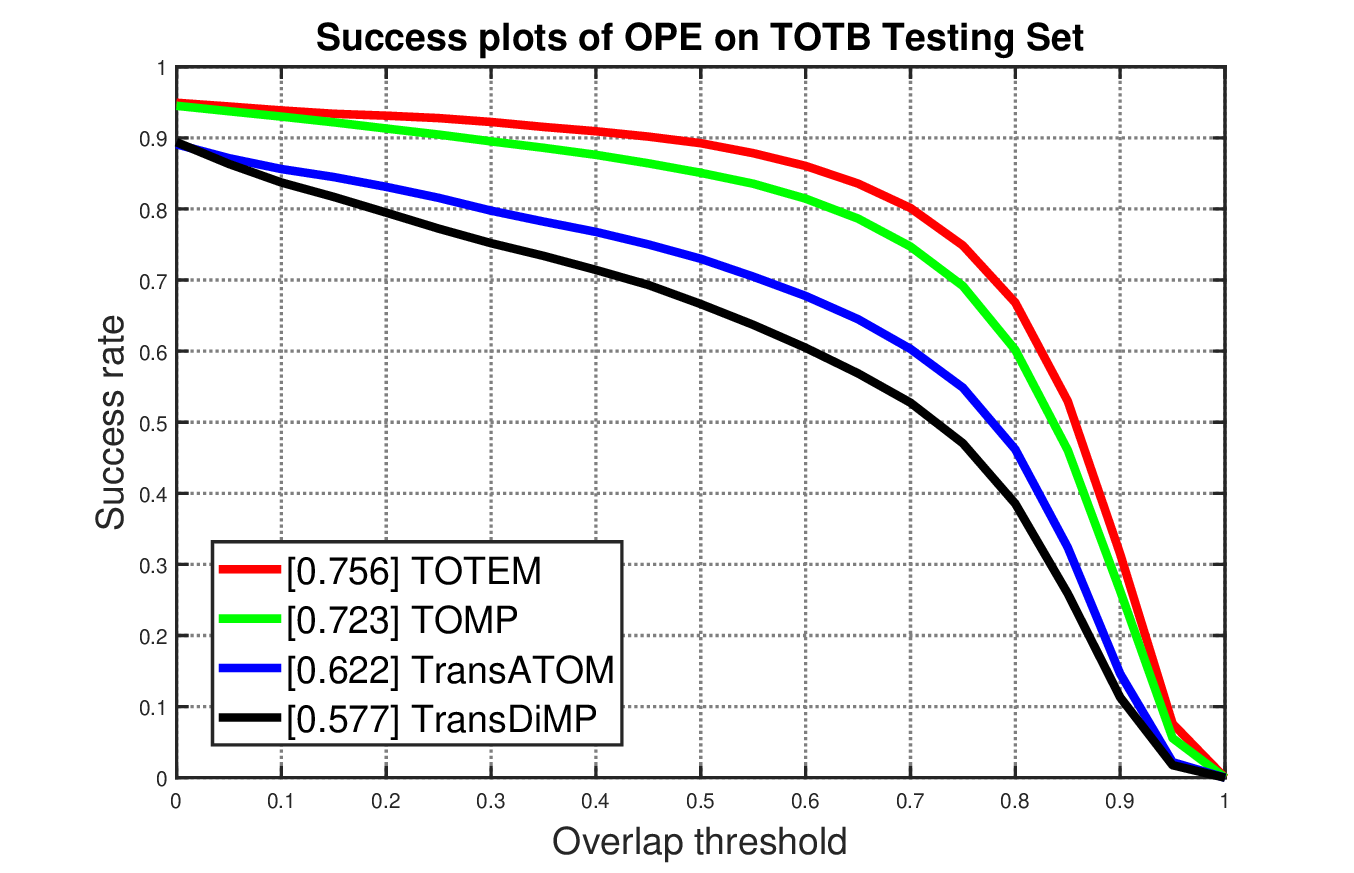}
            \caption[Success Plot]%
            {{\small Success Plot}}    
            \label{fig:sotacmpsuc}
        \end{subfigure}
        \hfill
        \begin{subfigure}{0.327\textwidth}  
            \centering 
            \includegraphics[width=\linewidth]{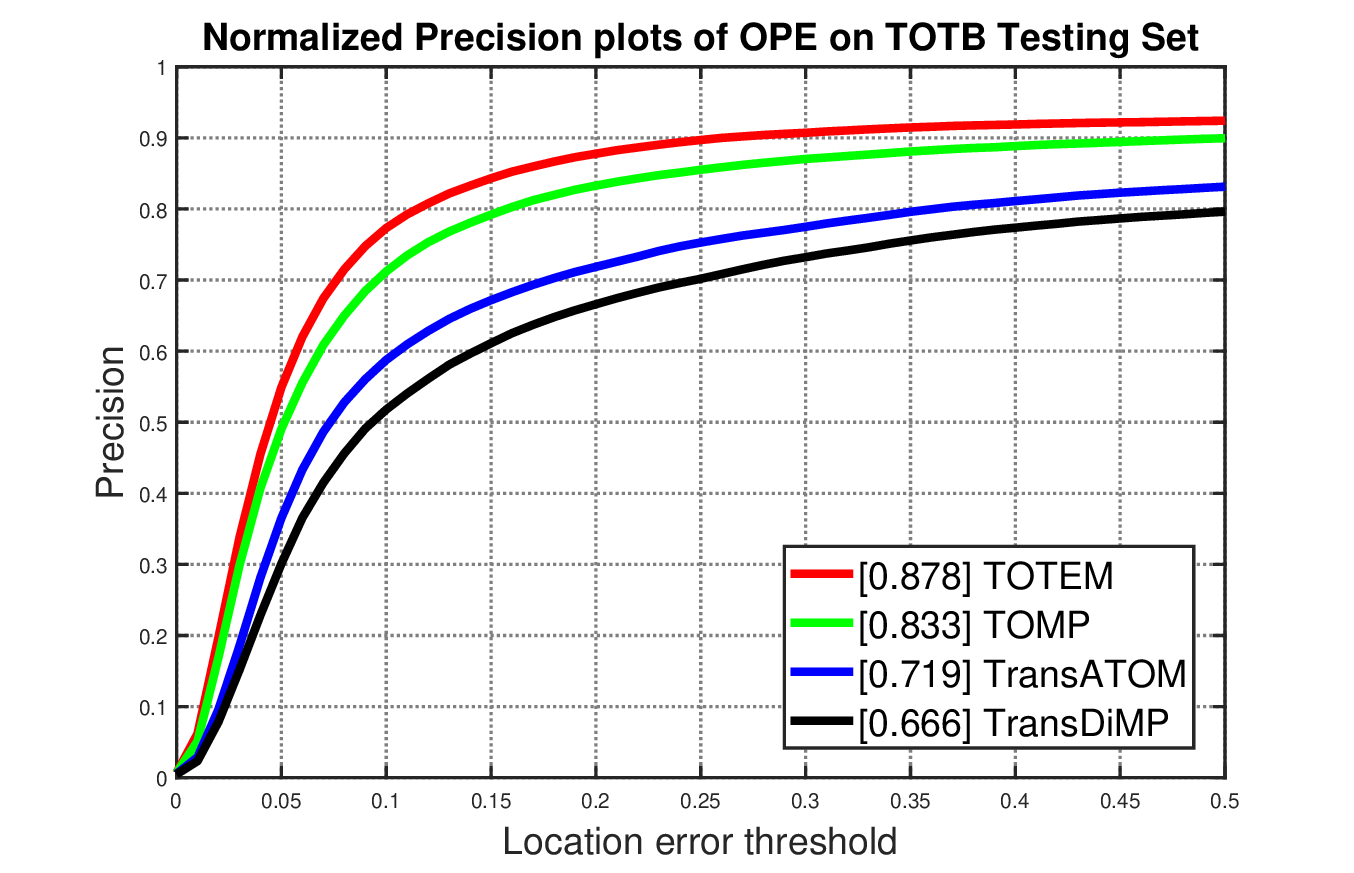}
            \caption[Normalized Precision Error Plot]%
            {{\small Normalized Precision Error Plot}}    
            \label{fig:sotacmpnpe}
        \end{subfigure}
        \hfill
        \begin{subfigure}{0.327\textwidth}  
            \centering 
            \includegraphics[width=\linewidth]{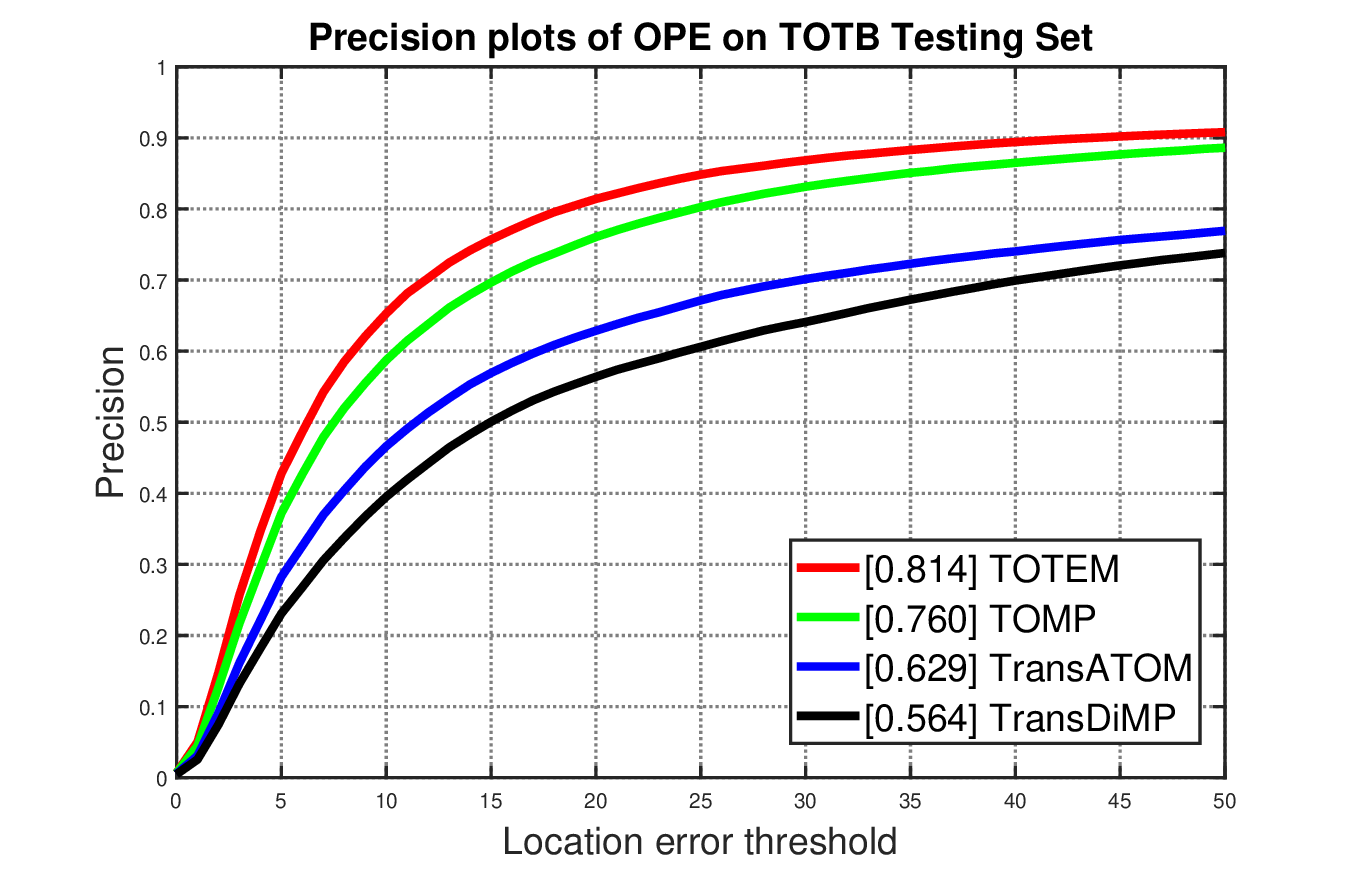}
            \caption[ Precision Error Plot]%
            {{\small  Precision Error Plot}}          
            \label{fig:sotacmppe}
        \end{subfigure}
        \caption[ SOTA Comparison]
        {\small Tracking performance of {\netname}, its baseline TOMP, and the two state-of-the-art trackers (TransATOM and TransDiMP),  in terms of precision, normalized precision and success metrics. Our tracker {\netname} achieves the best results with all three metrics. (Legend shows values in rate unit)
        } 
        \label{fig:sotacmp}
    \end{figure*}

\vspace{0mm}\noindent \textbf{Datasets.} For evaluation, we mainly use the Transparent Object Tracking Benchmark (TOTB) \cite{totb} dataset. This dataset comprises 15 common transparent object classes, with each containing 15 sequences (225 in total).
 Given the lack of any labeled training data sets, we split TOTB into two sections: a small section comprising 45 video sequences belonging to 3 object classes (\textit{Beaker, GlassBall, and WubbleBubble}) is used for training while the remaining 180 sequences (belonging to other 12 object classes) are used for testing. Since the fusion module is a lightweight module with only around 5 million learnable parameters, it can be trained from very little data. Additionally, the TOMP network is pretrained on a mix of LaSOT, GOT-10k \cite{got10k}, TrackingNet \cite{trackingnet}, and CoCo \cite{coco} datasets, while the Trans2Seg network (used as a transparency feature extractor) is trained over the Trans10k-v2 \cite{trans2seg, translab} dataset. 

\vspace{1mm} \noindent \textbf{Training.} 
TOMP and Trans2Seg are trained separately, and the learned weights are used for initializing the {\netname}. The weights of the fusion component are initialized with random Xavier initialization. The forward pass is performed on a sample containing a set of three input frames (two train frames and one test frame), each of size \(288 \times 288\). The classification and regression losses (\(L_1 \& L_2\)) are computed from the output of respective heads, and the back-propagation is performed to update the weights of the fusion module (with all other parameters frozen). 

We train our tracker on above mentioned splits of TOTB and LaSOT datasets for 25 epochs with 4000 image triplets sampled at every epoch. We set batch size as 18 and used ADAMW \cite{adamw} optimizer with a learning rate of 0.0001. The proposed fusion module uses a 4-layer transformer encoder and operates on 256-dim feature vectors. The training is performed in two steps, as described in section \ref{sec3:training}. We used 12GB TITAN Xp GPUs to train and test our model. Our model TOTEM runs at 6FPS during inference.

\subsection{Comparison study}
\label{sec4:sotacmp}

To establish baselines for comparison, we employ the recent transparent object tracker TransATOM \cite{totb} and its base ATOM \cite{atom}.  In addition, we also compare against TOTEM's base tracker TOMP~\cite{tomp}. To ensure fair comparison all models are fine-tuned end-to-end on the same training dataset. Note that, TOTEM is also fine-tuned end-to-end for this experiment.    


We report our results using the success (SUC), normalized precision (NPRE), and precision (PRE) metrics. The SUC plot in Fig. \ref{fig:sotacmpsuc} displays the overlap precision \(OP\) as a function of the threshold \(Th\). The NPRE and PRE plots (shown in Fig. \ref{fig:sotacmpnpe} and Fig. \ref{fig:sotacmppe} respectively) show the corresponding precision values against the overlap threshold. Trackers are ranked according to their area-under-the-curve (AUC) score for each plot, which is presented in the legend. Our proposed {\netname} tracker outperforms the previous state-of-the-art TransATOM tracker by a significant margin of 13.4\% in terms of Success AUC. Importantly, our proposed tracker outperforms its baseline TOMP by 3.3\% thanks to the transparency cues incorporated by our fusion module.

\subsection{Attribute analysis}
\label{sec4:attb}
\begin{figure*}[t!]
        \centering
        \begin{subfigure}{0.327\textwidth}
            \centering
            \includegraphics[width=\linewidth]{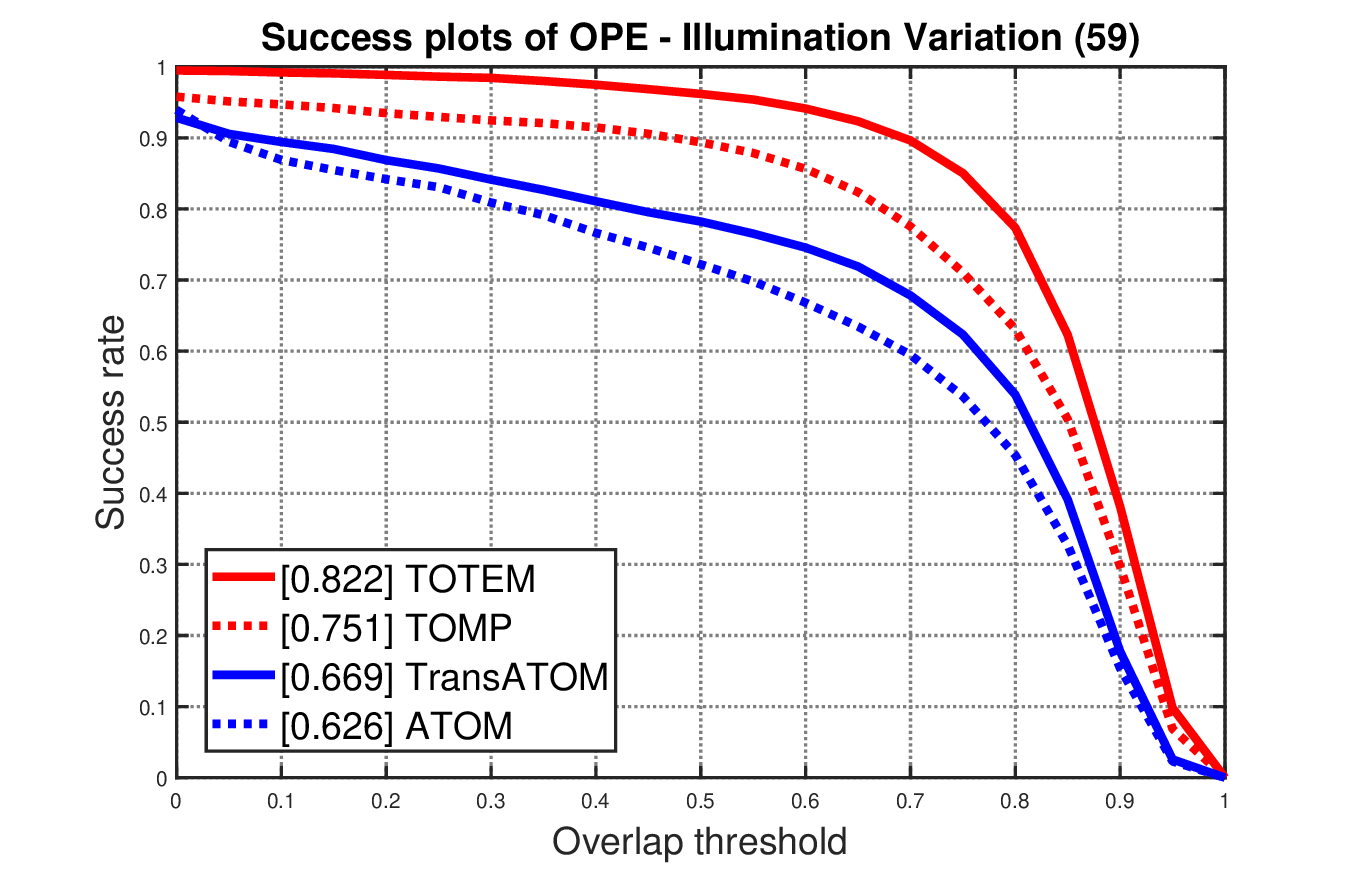}
            \caption[Illumination Variation]%
            {{\small Illumination Variation}}    
            \label{fig:attbposiv}
        \end{subfigure}
        \hfill
        \begin{subfigure}{0.327\textwidth}  
            \centering 
            \includegraphics[width=\linewidth]{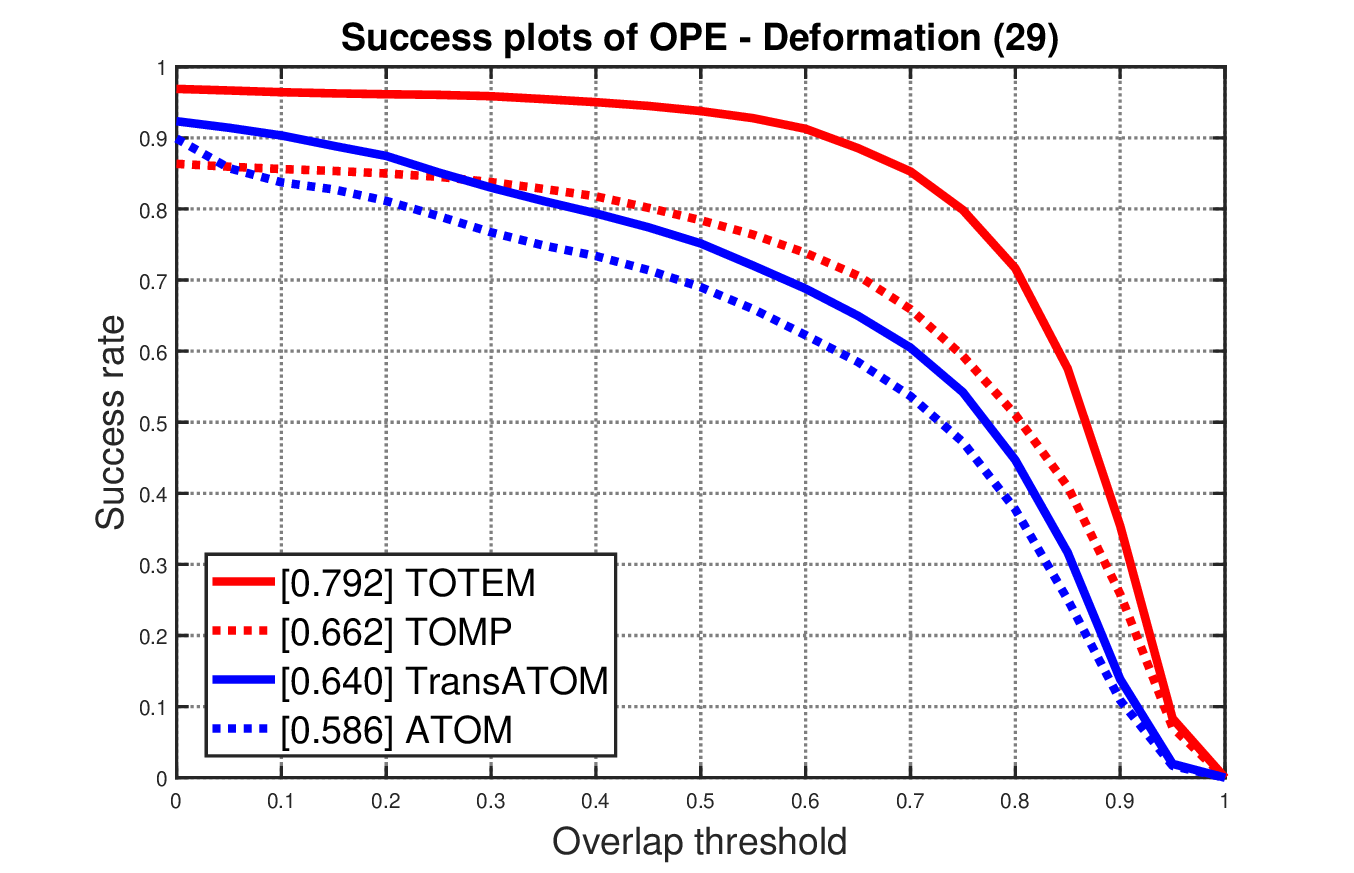}
            \caption[Deformation]%
            {{\small Deformation}}    
            \label{fig:attbposdef}
        \end{subfigure}
        \hfill
        \begin{subfigure}{0.327\textwidth}  
            \centering 
            \includegraphics[width=\linewidth]{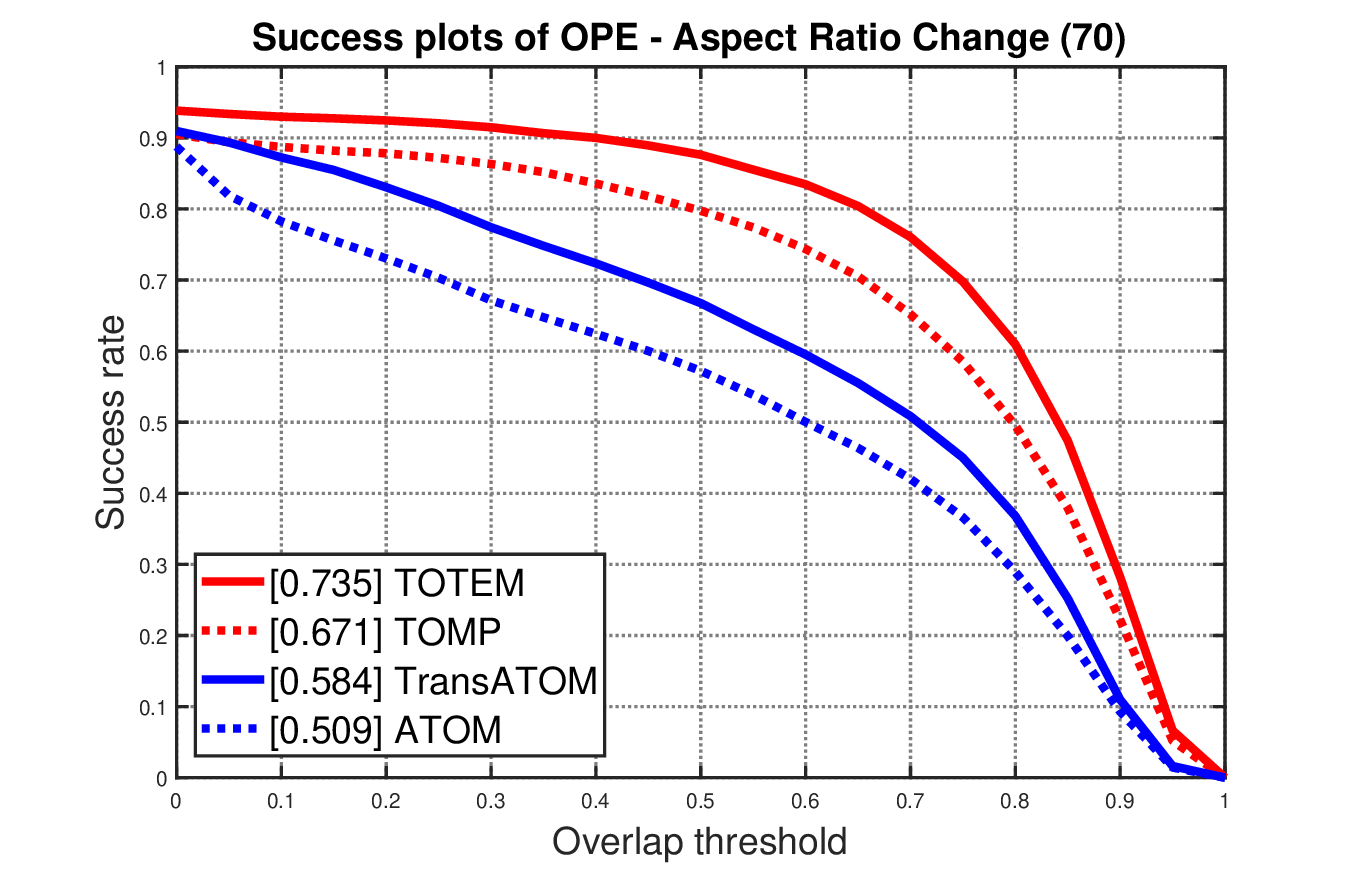}
            \caption[ Aspect Ratio Change]%
            {{\small  Aspect Ratio Change }}          
            \label{fig:attbposarc}
        \end{subfigure}
        \caption[  Analysis over tracking challenges: Illumination Variation, Deformation, Aspect Ration Change and Low Resolution   ]
        {\small Tracking performance of different tracking algorithms over the attribute on which TOTEM shows significant improvement over the baseline in terms of success metric. (Legend shows values in rate unit)
        } 
        \label{fig:attbpos}
    \end{figure*}
\begin{table*}
  \mbox{}\hfill
  \setlength\tabcolsep{6pt}
    \centering{\small
    \begin{tabular}{lccccccccccccc}
            \toprule
                      & All  &  IV  &  POC & DEF  &  MB  &  ROT &  BC  &  SV  &  FOC &  FM  &  OV  &  LR  & ARC \\ \midrule
            TOMP      & 70.0 & 75.1 & 67.1 & 66.2 & 65.5 & 70.3 & 68.1 & 68.3 & \textbf{46.4} & 59.9 & \textbf{61.4} & 63.1 & 67.1\\
            TransATOM & 62.2 & 66.9 & 58.0 & 64.0 & 52.5 & 60.0 & 59.6 & 59.6 & 27.6 & 50.0 & 36.5 & 58.3 & 58.4\\
            ATOM      & 58.6 & 62.6 & 55.6 & 58.6 & 50.0 & 55.3 & 58.1 & 53.8 & 31.6 & 45.4 & 40.0 & 58.7 & 50.9\\
            TOTEM(Ours)     & \textbf{75.6} & \textbf{82.2} & \textbf{71.4} & \textbf{79.2} & \textbf{69.5} & \textbf{73.8} & \textbf{69.4} & \textbf{74.5} & 41.1 & \textbf{68.1} & 60.9 & \textbf{74.0} & \textbf{73.5}\\
            \bottomrule
        \end{tabular}}

        \caption{\small Per attribute analysis on TOTB Test split. Value in each cell corresponds to success AUC metric (in \%) corresponding to the tracker and attribute. TOTEM scores the best against all the tracking challenges except \textit{Full-Occlusion} and \textit{Out-of-View}}
        \label{table:one}
\end{table*}

To analyze the performance of our algorithm on certain tracking challenges, we evaluate our tracker {\netname} under 12 different attributes. We explore the performance gain specifically due to the addition of transparency feature fusion by comparing {\netname} against the baseline TOMP. Also, we include the evaluations of ATOM vs TransATOM in this section so that we can compare the performance gains due to transparency feature infusion in our work against that in \cite{totb}.

Both the baselines TOMP and ATOM are directly adapted from their respective works \cite{tomp}, and \cite{atom} without any modifications, whereas TransATOM and {\netname} follow the same training settings as described in the above section \ref{sec4:sotacmp}.

Tab. \ref{table:one} lists the comparison results against all 12 attributes using the success AUC metric. We observe that {\netname} performs best on 10 out of 12 attributes. {\netname} shows a major improvement in the case of \textit{Illumination Variation}, \textit{Deformation}, \textit{Aspect Ration Change}, and \textit{Low Resolution} attributes (see Fig. \ref{fig:attbposiv}, \ref{fig:attbposdef} and \ref{fig:attbposarc} respectively) outperforming its baseline with Sucess AUC scores of 82.2\%, 79.2\%, 73.5\% and 74.0\% by 7.1\%, 13\%, 6.4\% and 10.9\% respectively. This huge improvement in tracking accuracy can be directly attributed to using transparency features in the pipeline. Deformation and aspect ratio changes are a result of variations in the target object's shape. Such variations are hard to deal with if a tracker cannot fully understand the target's appearance. For example, a backbone network that does not understand a target might encode two variant poses of it into embeddings that do not relate well. Such inefficiency in the backbone can further cause the model prediction module to perform poorly at generating kernel weights that produce accurate localization. In the case of the TOTEM tracker, the Trans2Seg model is extensively trained to understand transparent objects and thus has the ability to extract relevant features. For example, it might produce embeddings invariant to background patterns, given that such property benefits the network for performing segmentation tasks on transparent objects. Having transparency features fused into our baseline tracker's pipeline will directly help with better localization. In our case, the transparency features helped the model to perform better in case of appearance-varying situations.   



\subsection{Ablation Study}
\label{sec4:ablate}
Our tracker {\netname} benefits from three main components. First, we use TOMP as the baseline, which has a significant performance advantage over the other baselines (ATOM and DiMP, for example). Second, we utilize the Trans2Seg backbone along with its encoder to extract transparency features. Third, our proposed fusion module combines the transparency features into our baseline tracker's pipeline. In this section, we ablate each component and show that the design helps improve accuracy. We additionally evaluate our two-step training strategy against other methods. 

\vspace{1mm} \noindent \textbf{Baseline.}
\begin{table}
    \centering
      \setlength\tabcolsep{6pt}

    \begin{tabular}{lccc}
            \toprule
            Baseline Tracker                                & SUC   &  PE  & NPE \\ \midrule
            
            TOMP                          & \textbf{70.0}      & \textbf{72.2}          & \textbf{80.7} \\ 
            ATOM                          & 58.6      & 59.0          & 68.4 \\ 
            DiMP                          & 56.5      & 54.5          & 62.7 \\ \midrule
            TransATOM                     & 61.4 & 61.7 & 71.4 \\
            
            \bottomrule
            \end{tabular}

            \caption{\small Analysis of different backbones for tracking performance on TOTB using SUC score.}
        \label{table:two}
            
\end{table}
\label{sec4:ablatebaseline}
We evaluate our baseline model TOMP against the baselines of the other transparent object trackers as shown in Tab. \ref{table:two}. All the trackers follow their original configuration and are not pre-trained on TOTB. This analysis provides us with the portion of improvement we solely gain by using a transformer-based model predictor, independent of other factors. We observe that TOMP outperforms ATOM and DiMP by 11.4\% and 13.5\% in success AUC scores, respectively. TOMP provided a better starting point which in itself has surpassed the previous state-of-the-art tracker TransATOM by a margin of  8.6\%.

\vspace{1mm} \noindent \textbf{Transparency Features.}
\label{sec4:ablatetransfeat}
\begin{table}
  \setlength\tabcolsep{6pt}
    \centering{
        \begin{tabular}{lccc}
            \toprule
                                            & SUC   &  PE  & NPE \\ \midrule
            
            TOTEM-TE                           & 70.4      & 72.9          & 81.3 \\
            TOTEM-T                               & 70.3      & 72.7          & 81.8\\
            TOTEM             & \textbf{75.6}      & \textbf{81.4}          & \textbf{87.8}\\

            \bottomrule
            \end{tabular}%
    }

    \caption{\small Analysis of transparency features on tracking performance. }
    \label{table:ablatetransfeat}%
  
\end{table}

In this subsection, we ablate the components of Trans2Seg from TOTEM to analyze the benefit due to transparency features in our pipeline. 

We created a new tracker model, TOTEM-T, to enable a fair ablation study of transparency features. TOTEM-T uses our fusion module just like TOTEM, but it does not have transparency features in the input. This way, the only difference between TOTEM and TOTEM-T is the use of transparency features with the fusion module. Tab. \ref{table:ablatetransfeat} shows success (SUC) AUC results comparing TOTEM-T with TOTEM. TOTEM shows better accuracy in tracking with SUC, PRE, and NPRE metrics at 75.6\%, 81.4\%, and 87.8\%, respectively (with gains of 5.3\%, 8.7\%, and 6\%) compared to TOTEM-Ts 70.3\%, 72.7\%, and 81.8\%. This proves that transparency features are certainly beneficial. Further, we perform another ablation study with TOTEM-TE by removing the Transformer encoder component from the transparency feature extractor. This model is observed to perform fairly in comparison with TOTEM-T, signifying that the encoder block from Trans2Seg plays a crucial role in providing transparency awareness.


\vspace{1mm} \noindent \textbf{Fusion Module.}
\label{sec4:ablatefusionmodule}
\begin{table}
  \setlength\tabcolsep{6pt}
    \centering
    \begin{tabular}{lccc}
    \toprule
    Fusion Approach                               & SUC   &  PE  & NPE \\ \midrule
    TOTEM-MLPHead    & 66.3      & 67.8 & 77.4 \\
    TOTEM-\(e_\mathrm{query}\)   & 69.7      &  \textbf{73.1}             & 81.4\\
    TOTEM-FFNFuse             & 67.7      &  69.3                 & 79.4\\
    TOTEM                      & \textbf{70.2}      &  72.9                 & \textbf{82.4}\\
    \bottomrule
    \end{tabular}%

    \caption{\small Analysis of fusion module on tracking performance. }
    \label{table:ablatefusionmodule}%
\end{table}

We evaluate the effectiveness of the proposed fusion module by comparing it against the standard feed-forward network (FFN) based fusion. We create a model TOTEM-FFNFuse that uses an 8-layer feed-forward network that projects a concatenated feature (dimensions 512 = 256 + 256) into a fused feature (size 256dim). For fairness, we designed the FFN fusion module to have the same number of learnable parameters as our transformer-based fusion module. 
In Tab. \ref{table:ablatefusionmodule}, we observe that  TOTEM outperforms TOTEM-FFNFuse in the SUC metric by a margin of 2.5\%. This indicates that our transformer-based fusion module is effective in fusing the transparency features into the TOMP pipeline. 

Further, we ablate components within the fusion module. We first investigate the benefit of having a learnable query embedding \(e_\mathrm{query}\) in the fusion stream by comparing TOTEM with an ablated variant, TOTEM-\(e_\mathrm{query}\), that lacks a learnable feature in its fusion input. Tab. \ref{table:ablatefusionmodule} shows TOTEM-\(e_\mathrm{query}\) has a slight performance drop of .5\% and 1.0\% in SUC and NPRE metrics, respectively, while showing only a 0.2\% improvement in PE metrics. Overall a slight improvement is noticed. Given that \(e_\mathrm{query}\) is only a 256-sized floating point weight and has comparably less computation overhead, the design choice of including it is beneficial.   

We also ablate the MLP module \(\phi\) that projects the fused features into the encoder input space. In this test, we create a variant TOTEM-\(\phi\) by ablating the MLP. In Tab. \ref{table:ablatefusionmodule}, when compared to TOTEM this variant showed a significant performance drop of 3.9\% in the SUC metric, indicating that the MLP projection module is crucial to the performance of fusion. 

\vspace{1mm}\noindent \textbf{Two-step training approach}
\label{sec4:ablatetwosteptrain}
\begin{table}
  \setlength\tabcolsep{6pt}

    \centering
        \begin{tabular}{lccc}
        \toprule
        Training Method                                & SUC   &  PE  & NPE \\ \midrule
        One step train                          & 70.2      & 72.9          & 82.4 \\
        Two step train                          & 71.3      & 74.7          & 83.0 \\  \midrule
        Two step train + end-to-end fine-tune             & \textbf{75.6}      & \textbf{81.4}          & \textbf{87.8} \\
        
        \bottomrule
        \end{tabular}

    \caption{\small Analysis of training approach on tracking performance. }
    \label{table:ablatetwosteptrain}%
\end{table}

Along with the new fusion module, we proposed a two-step approach for training it, reasoning that it helps the module use transparency features well. In Tab. \ref{table:ablatetwosteptrain}, we produced results comparing the two-step approach with simple one-step training. Here, we observe that the two-step approach outperforms the simple method by 1.1\% in the SUC metric. We also notice 1.8\% and 0.5\% gains with our approach in the PRE and NPRE metrics, respectively. 

Additionally, we demonstrate the efficiency of end-to-end fine-tuning when performed in complement with two-step training. Here, we fine-tune our entire tracker instead of only updating the fusion module's weights. With this extra tuning, TOTEM observes a performance improvement of 4.3\% in the SUC metric. Interestingly, observing from the SUC metrics of TOMP from Fig. \ref{fig:sotacmpsuc} and Tab. \ref{table:two}, we only observe a gain of 2.3\% with the baseline. This further shows the benefit of fusing the transparency features.



\section{Conclusion}
\label{chap:conclusion}

In this work, we explored an important yet under-explored problem of \textit{transparent object tracking}. We proposed a novel tracker architecture named TOTEM, which benefits from understanding the unique texture properties of transparent objects. In particular, we successfully transferred the information learned from \textit{transparent object segmentation} to \textit{tracking} by using the pretrained Trans2Seg (a segmentation network) model to aid our tracker with extra transparency cues. In addition, we presented a new fusion module that learns to fuse features from different streams and projects them to the feature space of the original stream. Due to the projection property, our module can be added/removed from the tracker pipeline without retraining the network. Further, we explored a new training strategy \textit{i.e., two-step training} that explicitly improves the fusion performance of our proposed module. Comprehensive experiments are performed, showing that TOTEM considerably outperforms the previous state-of-the-art and its baseline. Our ablation studies show that each design choice we made toward TOTEM has a positive contribution to its performance.

\vspace{1mm} \noindent \textbf{Future Work.} The fusion module combined with our two-
step training strategy shows promising performance gains. In the future, we would extend the module to aid generic trackers in gaining application-specific skills. For example, camouflaged object tracking can be made possible without explicit training data with the help of our fusion techniques.

\section*{Acknowledgement}
We thank all reviewers for valuable comments and suggestions. The work was supported in part by US National Science Foundation Grants 2006665, 2128350, and 2128187. This work is also supported in part by by Air Force Office of Scientific Research FA 9550-23-2-0002.  The support of these agencies is gratefully acknowledged. Any opinions, findings, and
conclusions, or recommendations expressed in this material are those of the authors and do not necessarily reflect the views of
the National Science Foundation or the United States Air Force.


{
\bibliographystyle{IEEEtran}
\bibliography{egbib}
}

\end{document}